%% file: 0_main.tex
\documentclass[11pt]{article}

\usepackage[final]{acl}

\usepackage{times}
\usepackage{latexsym}
\usepackage{booktabs}

\usepackage[T1]{fontenc}
\usepackage{amsmath}
\usepackage{multirow}

\usepackage[utf8]{inputenc}

\usepackage{microtype}

\usepackage{inconsolata}

\usepackage{graphicx}
\usepackage{enumitem} 

\usepackage{tcolorbox}
\tcbuselibrary{skins, breakable}
\usepackage{xcolor}

\usepackage{hyperref}

\newtcolorbox{promptbox}[2][]{
  colback=gray!5!white,    
  colframe=gray!75!black,    
  title=\textbf{#2},        
  fonttitle=\small\sffamily,
  fontupper=\small\ttfamily, 
  boxrule=0.8pt,
  sharp corners,         
  breakable, 
  enhanced,
  left=5pt, right=5pt, top=5pt, bottom=5pt,
  #1
}
\title{If Only My CGM Could Speak: A Privacy-Preserving Agent for Question Answering over Continuous Glucose Data} 

\author{
  \textbf{Yanjun Cui\textsuperscript{1}},
  \textbf{Ali Emami\textsuperscript{2}},
  \textbf{Temiloluwa Prioleau\textsuperscript{2}}\thanks{Co-corresponding authors.},
  \textbf{Nikhil Singh\textsuperscript{1}}\footnotemark[1]
\\
  \textsuperscript{1}Dartmouth College \quad
  \textsuperscript{2}Emory University
\\
  \texttt{
    \href{yanjun.cui.gr@dartmouth.edu}{yanjun.cui.gr@dartmouth.edu},
    \href{ali.emami@emory.edu}{ali.emami@emory.edu}, }\\
  \texttt{
    \href{temiloluwa.prioleau@emory.edu}{temiloluwa.prioleau@emory.edu}, 
    \href{nikhil.u.singh@dartmouth.edu}{nikhil.u.singh@dartmouth.edu}
  }
}

\begin{document}
\maketitle
\begin{abstract}
Continuous glucose monitors (CGMs) used in diabetes care collect rich personal health data that could improve day-to-day self-management. However, current patient platforms only offer static summaries which do not support inquisitive user queries. Large language models (LLMs) could enable free-form inquiries about continuous glucose data, but deploying them over sensitive health records raises privacy and accuracy concerns. In this paper, we present \textbf{CGM-Agent}, a privacy-preserving framework for question answering over personal glucose data. In our design, the LLM serves purely as a reasoning engine that selects analytical functions. All computation occurs locally, and personal health data never leaves the user's device. For evaluation, we construct a benchmark of 4,180 questions combining parameterized question templates with real user queries and ground truth derived from deterministic program execution. Evaluating 6 leading LLMs, we find that top models achieve 94\% value accuracy on synthetic queries and 88\% on ambiguous real-world queries. Errors stem primarily from intent and temporal ambiguity rather than computational failures. Additionally, lightweight models achieve competitive performance in our agent design, suggesting opportunities for low-cost deployment. We release our code and benchmark to support future work on trustworthy health agents.\footnote{Code available at: \url{https://github.com/yanjunCC/cgm-agent-release}}
\end{abstract}
\input{1_introduction}

\input{2_related_work}
\input{3_dataset}
\input{4_cgm_agent}
\input{5_experiment}
\input{6_conclusion}

\input{7_limitations_and_ethics}

\bibliography{custom}

\newpage
\appendix
\input{8_appendix}

\end{document}

%% file: 1_introduction.tex
\begin{figure*}[!htb]
    \centering
    \includegraphics[width=\linewidth]{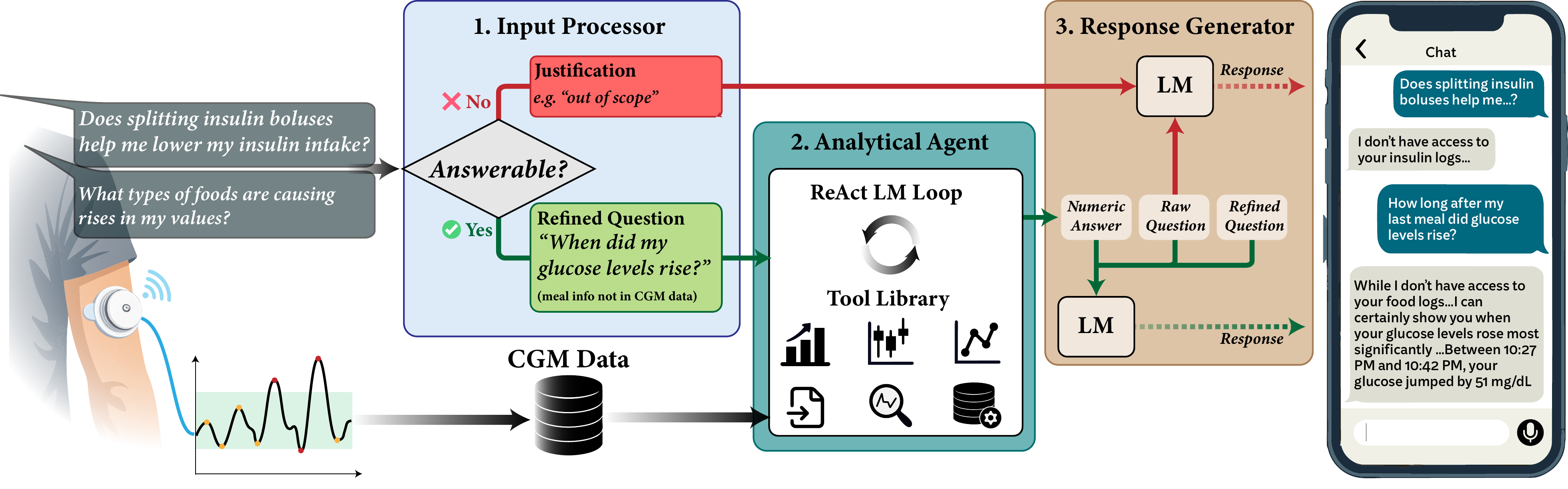}
    \caption{Overview of \textbf{CGM-Agent}. The pipeline consists of three layers: (1) the \textit{Input Processor} resolves ambiguous queries, (2) the \textit{Analytical Agent} generates tool calls executed in a local sandbox, and (3) the \textit{Response Generator} synthesizes a natural language answer. Raw CGM data (shown in the bottom sandbox) never crosses the privacy boundary; only function calls and aggregated metrics are exchanged with the LLM.}
    \label{fig:banner}
\end{figure*}

\section{Introduction}

Modern health sensors generate dense longitudinal records that, in principle, support fine-grained understanding of personal physiology and behavior. In diabetes care, continuous glucose monitors (CGMs) collect data for close monitoring of glucose patterns roughly every five minutes~\citep{american20252, dexcom_g6, abbott_freestyle_libre3}. Over months of CGM use, these devices accumulate detailed records of glucose fluctuations, encoding behavioral and physiological signals that could inform lifestyle decisions, medication adjustments, and clinical conversations. However, current user-interaction platforms like Dexcom Clarity and LibreView offer only static, pre-defined summaries \citep{dexcom2025official, libreview2024report}, thereby limiting the ability for a user to explore and learn hidden insights from their own data. For example, when a user has targeted questions like,\textit{``How does my glucose control on weekdays compare to weekends?''}, they have to download their data and manually compute the needed metrics to find answers to their questions. The gap between the \textit{richness} of CGM data and \textit{rigidity} of existing interfaces leaves users without personalized, on-demand insights they need to maximally benefit from their personal health data.

LLMs offer a promising solution for a conversational interface through which users can ask targeted questions in plain language. However, deploying LLMs over sensitive health data raises serious concerns. Cloud-based models would require transmitting personal health data to external servers which introduces significant safety and privacy risks. Moreover, LLMs are prone to numerical errors when processing time-series data, making them less reliable for direct use \citep{healey2024llm, healey2025case}.

In this paper, we present \textbf{CGM-Agent}, a modular framework that enables free-form question answering (QA) on personal glucose data while ensuring that personal health data never leaves the user's device (Figure~\ref{fig:banner}). Our key approach is to decouple \textit{reasoning} from \textit{computation} so that the LLM serves \textit{only} as a reasoning engine that interprets user intent and selects appropriate analytical functions, while all data processing occurs in a local execution sandbox. In our implementation, the LLM receives only aggregated CGM metrics, never raw time-series, eliminating the need to transmit sensitive personal health data to the cloud.

Our main contributions are:

\paragraph{A Grounded QA Benchmark.} We construct a dataset of 4,180 questions spanning both synthetic templates grounded in clinical guidelines \citep{battelino2019clinical, american20256} and real user queries sourced via interviews with CGM users. All ground-truth answers are derived from deterministic execution of a custom-designed Python toolkit developed in this work, ensuring reproducibility (\S\ref{sec:dataset}).

\paragraph{A Privacy-Preserving Agent Architecture.} We propose a three-layer pipeline that separates LLM reasoning from data access. Confining raw CGM data to a local sandbox and exposing only function calls to the model supports accurate, personalized responses without compromising user privacy (\S\ref{sec:agent}).
    
\paragraph{Empirical Analysis of Tool-Augmented LLMs.} We benchmark 6 leading proprietary and open-weight models through a layer-wise evaluation. Our results show that lightweight models achieve competitive accuracy on well-structured queries, while real-world, often ambiguous user questions expose significant challenges in intent disambiguation and temporal grounding (\S\ref{sec:experiments}).

%% file: 2_related_work.tex
\section{Related Work}

\paragraph{Tool-Augmented LLM Reasoning.}
Large language models are prone to errors in mathematical reasoning and factual recall, motivating work on tool augmentation \citep{schick2023toolformer, yao2022react}. Toolformer \citep{schick2023toolformer} showed that LLMs can learn to invoke external APIs via self-supervised learning; ReAct \citep{yao2022react} established interleaving reasoning traces with executable actions for multi-step problem solving. These frameworks have been extended to domain-specific settings: SciAgent \citep{ma2024sciagent} and ToolHop \citep{ye2025toolhop} advance tool-augmented reasoning and evaluation in scientific domains, while MMedAgent \citep{li2024mmedagent} and MeNTi \citep{zhu2025menti} adapt tool calling to clinical decision support. However, these systems typically operate over structured databases or static documents rather than personal longitudinal data, and do not address the privacy constraints inherent in patient-facing applications.

\paragraph{LLMs for Diabetes Management.}
Prior work has applied LLMs to diabetes education, coaching, and forecasting, yet evaluations have exposed critical limitations \citep{abbasian2024knowledge, mitchell2025t2, mamun2025llm, li2025llm}. For example, \citet{healey2024llm, healey2025case} showed that GPT-4 frequently misapplies diabetes-specific calculations and fails at CGM pattern recognition. DM-Bench \citep{cardei2025dm} benchmarks LLMs on patient-facing diabetes decision-making tasks across 15,000 individuals with multimodal behavioral logs. Furthermore, commercial systems such as ChatCGM \citep{ChatCGM2024Privacy} and Diabetes Cockpit \citep{CockpitApp2025}, which reflect growing market demand, offer LLM-powered glucose interfaces but remain closed-source with undisclosed evaluation procedures. Our work complements these efforts by providing an open benchmark with deterministic, reproducible ground truth, and a privacy-preserving architecture that keeps raw CGM data local.

\paragraph{Privacy in Personal Health AI.}
Cloud-based LLMs require transmitting sensitive records to external servers, raising concerns about data security and regulatory compliance \citep{winslow2025principle}. CGM data carries significant privacy risks, thus many existing platforms operate under regulatory gaps that leave patients with little to no control over how their data is collected, stored, or shared~\cite{downey2025urgent,randine2025privacy}. Some commercial systems mitigate this by sending only statistical summaries to the LLM, however this approach limits the richness of analysis and is proprietary in most cases \citep{CockpitApp2025}. Prior work on privacy-preserving LLM inference formalizes the \textit{Honest-but-Curious} (HbC) threat model for cloud providers \citep{li2024confidential}: the provider executes instructions faithfully but may log and exploit transmitted data. Locally deployed LLMs have been proposed as one response to this threat in healthcare \citep{wiest2024privacy,MONTAGNA2025100552}, but typically at the cost of model capability. In contrast, our work proposes an open, modular architecture in which raw CGM data never leaves the local device; the LLM receives only aggregated metrics, whose low temporal resolution provides protection against re-identification \citep{portability2012guidance,el2011systematic}. 

%% file: 3_dataset.tex
\section{Dataset} \label{sec:dataset}

We construct a QA benchmark for CGM data comprising 4,180 queries. The dataset combines synthetic templates grounded in clinical guidelines with real user questions collected through formative interviews with current CGM users. All ground-truth answers are derived from deterministic execution of a validated Python toolkit, ensuring mathematical precision and reproducibility.

\subsection{CGM Data Source} \label{sec:cgm-data-source}

For an empirical evaluation, we curate CGM records from 19 individuals drawn from two public datasets, covering both Type 1 Diabetes (T1D) and Type 2 Diabetes (T2D). We prioritized subjects with the longest recording duration, thereby selecting 11 T1D subjects from AZT1D \citep{khamesian2025type} (CGM duration: 45--49 days, 5-minute sampling interval), and 8 T2D subjects from ShanghaiT2DM \citep{zhao2023chinese} (CGM duration: 24--179 days, 15-minute sampling interval). The mapping between anonymous subject identifiers and original dataset identifiers is provided in Appendix~\ref{app:subject-selection}.

This combination provides a realistic testbed as AZT1D offers high-frequency, continuous recordings ideal for verifying the functional correctness, while ShanghaiT2DM introduces challenges of sparser sampling and significant missing data segments (up to 153 missing days for some subjects). Together, the combined dataset reflects the heterogeneity of real-world CGM use, spanning a wide range of Time in Range (TIR, 70–180 mg/dL) values (41.5–98.7\%) and missing data rates (Table~\ref{tab:subject_stats}).

\begin{table}[t]
\centering
\small
\begin{tabular}{l r r}
\toprule
\textbf{Characteristic} & \textbf{AZT1D} & \textbf{ShanghaiT2DM} \\
\midrule
Diabetes Type          & T1D         & T2D \\
Subjects               & 11          & 8 \\
Sampling Interval      & 5 min       & 15 min \\
Duration (days)        & 45--49      & 24--179 \\
Missing Days           & 0           & 0--153 \\
Missing Samples (\%)   & 1.2--2.4    & 4.0--86.2 \\
TIR Range (\%)         & 63.2--92.7  & 41.5--98.7 \\
\bottomrule
\end{tabular}
\caption{CGM dataset used in this study ($N = 19$).}
\label{tab:subject_stats}
\end{table}

\subsection{Analytical Toolkit} \label{sec:toolkit}

To ensure clinically meaningful and reproducible evaluation, we developed a Python toolkit that serves two purposes: (1) generating ground-truth answers for benchmark queries, and (2) providing executable functions for the agent at inference time. We organize the toolkit into two tiers.

\paragraph{Standard Glycemic Metrics.}
Informed by the international consensus recommendation report for CGM data interpretation \citep{battelino2019clinical} and the ADA Standards of Care \citep{american20256}, this tier implements standard CGM metrics such as mean glucose, Time in Range (TIR, 70--180 mg/dL), Time Below Range (TBR, $<$70 mg/dL), and Time Above Range (TAR, $>$180 mg/dL), while enabling statistical aggregation and comparative analysis across user-defined date ranges. For each specified time window, the system calculates these metrics alongside the CGM weartime during the associated date range. Following recommended guidelines, results are flagged when CGM weartime falls below 70\% \citep{battelino2019clinical}, enabling the agent to recognize when a query cannot be reliably answered due to insufficient glucose data.

\paragraph{User-Driven Analytics.}
Based on findings from our formative CGM user interviews (described below), we extended the toolkit to address user-expressed needs beyond standard metrics. This tier includes: (1) \textit{glucose excursion detection}, identifying rapid glucose rises or falls exceeding a threshold rate (e.g., $>$2 mg/dL/min), and (2) \textit{trend visualization}, generating plots of glucose trajectories over specified periods. Full function signatures are provided in Appendix~\ref{sec:appendix_toolkit}.

\subsection{Question Collection}

We constructed questions through two complementary approaches: synthetic generation from templates and collection from real users.

\paragraph{Synthetic Questions.}
We developed a question generation factory to systematically create diverse analytical queries: single-metric queries (\textit{``What was my TIR on [Date A]]?''}), and temporal comparisons (\textit{``Compare my average glucose between [Time Period A] and [Time Period B]''}). By programmatically instantiating templates with randomized time windows (e.g., single dates, duration, intraday) and injecting both valid CGM metrics and out-of-scope features (e.g., insulin, sleep), this engine produced 2,470 synthetic QA pairs using CGM data from 19 subjects as described in Section \ref{sec:cgm-data-source}; thereby ensuring robust test coverage. 

\paragraph{Real User Questions.}
To capture authentic information needs, we conducted semi-structured interviews with three individuals who have diabetes and actively use CGM devices (IRB-approved, Appendix~\ref{sec:appendix_recruitment}). Participants were asked to pose questions they would want to ask about their own glucose data, yielding 30 seed questions. While modest in scale, this formative study identified important gaps between clinical metrics and patient priorities. Users expressed strong interest in understanding glucose \textit{fluctuations} and \textit{trends}, not just aggregate statistics. For example, questions like \textit{``What are my typical 24-hour glucose patterns?''} require trend analysis rather than single-value metrics. These findings directly informed our Tier 2 toolkit extensions.

After de-identification and normalization of seed questions, we instantiated those questions through varying dates, time windows, and thresholds using CGM data from all 19 subjects described in Section \ref{sec:cgm-data-source} to generate 1,710 user-derived questions. 

\paragraph{Handling Unanswerable and Underspecified Questions.}
Real user questions often reference external context unavailable in CGM data alone (e.g., meals, exercise, insulin doses). We categorize these into two types:

\begin{itemize}[itemsep=2pt, parsep=0pt, topsep=0pt]
    \item \textbf{Unanswerable:} Questions requiring modalities outside CGM scope are labeled as unanswerable (e.g., \textit{``Does splitting boluses reduce my insulin intake?''}) . In this case, the agent should recognize and refuse these gracefully.
    
    \item \textbf{Proxy Questions:} Questions referencing unobserved behaviors that can be approximated via time windows (windows which could, for example, be obtained from the user via a dialog). For example, \textit{``How does exercise affect my TIR?''} cannot be answered without activity logs, but can be instantiated as \textit{``Compare TIR between 5--7 PM and other times today''} if the user typically exercises in that window. We manually mapped such questions to proxy instantiations, acknowledging that the agent provides \textit{approximate} analyses that may prompt user reflection rather than definitive answers.
\end{itemize}

The final benchmark dataset includes 513 unanswerable questions and 399 proxy questions to support evaluation of the agent's robustness to realistic edge cases. Detailed user questions and instantiation mappings are provided in Appendix~\ref{sec:appendix_questions}. Table~\ref{tab:qa_dataset_stats} summarizes the final dataset composition.

\begin{table}[!htb]
\centering
\small
\begin{tabular}{l l r}
\toprule
\textbf{Source} & \textbf{Category} & \textbf{Count} \\
\midrule
\multirow{1}{*}{Synthetic} & Template-generated & 2,470 \\
\midrule
\multirow{3}{*}{User-derived} & Directly answerable & 798 \\
& Proxy (instantiated) & 399 \\
& Unanswerable & 513 \\
\midrule
\multicolumn{2}{l}{\textbf{Total}} & \textbf{4,180} \\
\bottomrule
\end{tabular}
\caption{Composition of the CGM-QA benchmark. Proxy questions are user queries referencing unobserved behaviors, mapped to answerable approximations.}
\label{tab:qa_dataset_stats}
\end{table}

\subsection{Ground Truth Construction}

A distinguishing feature of our benchmark is that ground truth is defined as the \textit{output of deterministic program execution}, not human annotation. For each question, we specify the corresponding reference procedure of function call(s) and their associated parameters. The ground-truth answer is the numerical result of executing these functions on the subject's CGM data.

This design ensures: (1) \textbf{mathematical precision}---answers are computed, not estimated; (2) \textbf{reproducibility}---any system can verify answers by re-running the toolkit; and (3) \textbf{scalability}---similar new questions can be added with minimal manual annotation effort. This methodology establishes a rigorous standard for evaluating tool use in healthcare, where numerical accuracy is paramount.

%% file: 4_cgm_agent.tex
\section{Agent Architecture} \label{sec:agent}

Figure~\ref{fig:banner} illustrates the overall architecture of CGM-Agent. In this section, we describe each component in detail. We begin by stating our core design principles, then walk through the three layers of the pipeline: Input Processor, Analytical Agent, and Response Generator. 

\subsection{Design Principles}

Our architecture is guided by two principles:

\paragraph{Separation of Reasoning and Computation.}
The LLM serves purely as a reasoning engine: it interprets user intent, resolves ambiguities, selects appropriate analytical functions, and synthesizes natural language responses. All numerical computation over CGM data occurs in a local execution sandbox. To maximally preserve user privacy, the LLM never receives raw glucose readings in its context window, it only aggregates the results from function calls. 

\paragraph{Modularity for Evaluation.}
Each layer performs a distinct function. This separation enables targeted evaluation: we can measure whether the Input Processor correctly resolves temporal references, whether the Analytical Agent selects appropriate tools, and whether the Response Generator produces accurate and readable output. Thus, errors can be localized to specific components.

\subsection{Pipeline Overview}

Given a user query $q$, the pipeline proceeds as follows. The Input Processor $\mathcal{P}$ validates feasibility, resolves temporal references, and maps vague terms to precise system features, producing a refined query $q'$. The Analytical Agent $\mathcal{T}$ decomposes $q'$ into function calls and executes them locally with user's private CGM data $D$, producing numerical results $R$. The Response Generator $\mathcal{G}$ synthesizes a natural language response $y$:

\begin{equation}
q \;\xrightarrow{\;\mathcal{P}\;}\; q' \;\xrightarrow[\text{Tools}(D)]{\;\mathcal{T}\;}\; R \;\xrightarrow{\;\mathcal{G}(q, q', R)\;}\; y
\end{equation}

If the Input Processor determines that $q$ is unanswerable (e.g., requires insulin logs), it generates an explanation $e$ and the pipeline bypasses the Analytical Agent:
\begin{equation}
q \;\xrightarrow{\;\mathcal{P}\;}\; e \;\xrightarrow{\;\mathcal{G}(q, e)\;}\; y
\end{equation}

Appendix~\ref{sec:control_flow_appendix} illustrates these two paths with representative examples. 

\subsection{Layer 1: Input Processor}

The Input Processor serves as a gatekeeper, performing two functions before any data access.

\paragraph{Feasibility Classification.}
As mentioned previously, the LLM evaluates whether the query can be answered given available modalities (CGM only). 

\paragraph{Temporal and Feature Instantiation.}
For answerable queries, the LLM grounds user query $q$ into a fully specified format $q'$. This process involves resolving natural language time expressions (e.g., \textit{``last weekend''}) into precise datetime ranges, and normalizes vague descriptors into fully specified variables, transforming inputs like \textit{``how's my sugar''} into precise requests for \textit{``mean glucose''} and \textit{``data sufficiency''} (CGM weartime) compatible with the analytical toolkit.

\paragraph{Example.}
\begin{quote}
\small
\textbf{Input:} ``What's my TIR last weekend?'' \\
\textbf{Context:} Current date is 2024-01-10 (Wednesday) \\
\textbf{Output:} \texttt{is\_answerable: True,\\
refined\_question: What's my TIR and CGM weartime on 2024-01-06 and 2024-01-07?}
\end{quote}

\subsection{Layer 2: Analytical Agent}

The Analytical Agent translates the refined query $q'$ into executable function calls and retrieves numerical results. Crucially, the LLM interacts with CGM data only through predefined tool interfaces, never directly accessing raw glucose readings.

\paragraph{Planning and Decomposition.}
Some queries require multiple analytical steps. For example, \textit{``What's my TIR over this week and last week?''} involves computing TIR separately for two time windows. The LLM decomposes such queries into a sequence of atomic sub-tasks, each mapped to a single function call.

\paragraph{Tool Execution.}
For each sub-task, the LLM generates a function call specifying the operation and parameters (metric type, date range, reference thresholds). This call is dispatched to the local execution sandbox, where the Python toolkit (Section~\ref{sec:toolkit}) processes the user's CGM data and returns aggregated results. The LLM receives only these aggregated outputs (e.g., \texttt{\{avg\_TIR\_sufficient\_weartime: 72\%, days\_sufficient\_weartime: 4\}}), never the underlying time series.

\paragraph{Modularity and Extensibility.}
Given that the Analytical Agent relies entirely on deterministic tool calls rather than end-to-end parametric modeling, new data modalities can be integrated by registering a corresponding local executor, without modifying the overarching framework. 

\paragraph{Example.}
The following example illustrates the decomposition and execution process. Note that the toolkit computes TIR on a per-day basis and then aggregates across days, reporting separate statistics for days with sufficient CGM weartime ($\geq$70\%) and for all recorded days. 

\begin{quote}
\small
\textbf{Input:} $q'$ = What's my TIR over this week and last week? \\
\textbf{Decomposition:} \\
\quad Task A: What's my TIR and CGM weartime on (dates=this\_week)?\\
\quad Task B: What's my TIR and CGM weartime on (dates=last\_week)?\\
\textbf{Execution Results:} \\
\quad $R_A$ = \texttt{\{days\_sufficient\_weartime = 4, \\
avg\_TIR\_sufficient\_weartime: 72\%, \\
days\_all = 6, avg\_TIR\_all: 76\%} \\
\quad $R_B$ = \texttt{\texttt{\{days\_sufficient\_weartime = 5, \\
avg\_TIR\_sufficient\_weartime: 78\%, \\
days\_all = 7, avg\_TIR\_all: 86\%}}
\end{quote}

\subsection{Layer 3: Response Generator}
\label{sec:response_generator}
The final layer synthesizes the numerical results into a natural language response for the user.

\paragraph{Answerable Queries.}
For queries with valid results, the generator produces a response that: (1) directly answers the user's question with specific numbers, (2) explicitly states the time period analyzed and metrics instantiated to verify alignment with user intent, and (3) provides brief clinical context where appropriate (e.g., comparing TIR to recommended targets).

\paragraph{Unanswerable Queries.}
For queries flagged by Layer 1, the generator produces a response that: (1) acknowledges the user's question, (2) explains specifically why it cannot be answered with CGM data alone, and (3) suggests alternative approaches (e.g., consulting a healthcare provider or integrating additional data sources).

\paragraph{Example (Answerable).}
\begin{quote}
\small
\textbf{Input:} $q$ = ``How long after eating do my glucose levels rise?'' \\
\textbf{Results:} $R$ = \texttt{"(2021-08-29 09:37:00, 2021-08-29 09:52:00)": \\{"magnitude": 30.6, "speed": 2.04}}\\
\textbf{Response:} ``Based on your data from August 29th, a significant rise began at 9:37 AM, where your levels increased by 30.6 mg/dL in 15 minutes. Typically, glucose rises within 15--60 minutes after eating, depending on the food. Your levels were rising at about 2.0 mg/dL per minute. Tracking these patterns can help you understand how different meals affect you!''
\end{quote}

\paragraph{Example (Unanswerable).}
\begin{quote}
\small
\textbf{Input:} ``Does splitting insulin boluses help reduce my total daily insulin dose?'' \\
\textbf{Response:} ``I can't determine if splitting boluses reduces your total daily insulin, as I don't have access to your insulin logs. To answer this, we'd need to compare total units across different periods. I recommend reviewing your insulin records with your healthcare provider.''
\end{quote}

\vspace{0.5em}
\noindent All system prompts are provided in Appendix~\ref{sec:appendix_agent_prompts}.

%% file: 5_experiment.tex
\section{Experiments} \label{sec:experiments}

We evaluate CGM-Agent along two dimensions: (1) tool-use accuracy on synthetic queries with fully specified intent, and (2) end-to-end performance on ambiguous real-world user queries.

\subsection{Experimental Setup}

\paragraph{Models.}
We evaluate a total of 6 proprietary and open-weight LLMs as backbones for our agent:
\begin{itemize}[itemsep=2pt, parsep=0pt, topsep=0pt]
        \item Proprietary Models (Temp=1.0): GPT-5.2, GPT-5-Mini, Gemini 3.0 Pro and Flash.
    \item Open-Weight Models (Temp=0.6): Llama-4 Maverick (17B), Nemotron-Nano (9B).
\end{itemize}
All models are accessed via their respective APIs. We use the same system prompts across all models for consistency.

\paragraph{Evaluation Metrics.}
For the Analytical Agent (Layer 2), we report:
\begin{itemize}[itemsep=2pt, parsep=0pt, topsep=0pt]
    \item \textbf{Precision / Recall / F1:} Whether the agent selects the correct analytical function(s) and parameters. Precision measures the fraction of predicted function calls that match ground truth; Recall measures the fraction of ground-truth calls that were predicted.
    \item \textbf{Value Accuracy:} Whether the final numerical output matches the ground-truth value within a tolerance of $\pm 1\%$.
\end{itemize}
For the Input Processor (Layer 1), we report classification accuracy, precision, recall, and F1 for the feasibility decision (answerable vs.\ unanswerable).

\paragraph{Automated Evaluation.}
Given the scale of evaluation (4,180 queries), we use Gemini 3.0 Pro as an automated judge to assess function-call correctness and value matching. To validate this approach, we manually reviewed 40 randomly selected samples. The evaluation showed high alignment with human judgment, yielding a Mean Absolute Error (MAE) of $\approx 0.05$ for precision and $\approx 0.04$ for recall. Evaluation prompts are provided in Appendix~\ref{sec:appendix_eval_prompts}.

\subsection{Results on Synthetic Queries}

The synthetic dataset ($N=2,470$) serves as a controlled evaluation of the Analytical Agent (Layer 2), where temporal references and user intent are fully specified. This isolates tool-use capability from the ambiguity challenges present in real-world queries.

Table~\ref{tab:synthetic_results} presents the results. Key findings:

\begin{itemize}[itemsep=2pt, parsep=0pt, topsep=0pt]
    \item \textbf{Proprietary models achieve strong tool-use accuracy.} All proprietary models achieved 0.8+ F1 and three models have 0.94 Value Accuracy, demonstrating reliable function selection and parameter extraction.
    
    \item \textbf{Smaller models are competitive.} GPT-5-Mini and Gemini 3.0 Flash achieve Value Accuracy comparable to their larger counterparts ($\approx 0.94$), suggesting that tool-call does not require the largest-scale frontier models. This is encouraging for lower-cost deployments.
    
    \item \textbf{Open-weight models lag but remain viable.} Llama-4-17B achieves 0.75 Value Accuracy, trailing proprietary models but demonstrating that local-only deployments may be feasible with some performance trade-off.
\end{itemize}

\begin{table}[!htb]
\centering
\small
\renewcommand{\arraystretch}{1.15}
\begin{tabular}{l c c c c}
\toprule
\textbf{Model} & \textbf{Prec} & \textbf{Rec} & \textbf{F1} & \textbf{Val Acc} \\
\midrule
\multicolumn{5}{c}{\textit{Proprietary}} \\
\midrule
GPT-5.2 & 0.84 & 0.89 & \textbf{0.86} & 0.81 \\
GPT-5-Mini & 0.75 & 0.87 & 0.80 & \textbf{0.94} \\
Gemini 3.0 Pro & \textbf{0.92} & 0.71 & 0.80 & \textbf{0.94} \\
Gemini 3.0 Flash & 0.89 & 0.74 & 0.81 & \textbf{0.94} \\
\midrule
\multicolumn{5}{c}{\textit{Open-weight}} \\
\midrule
Llama-4-17B & 0.73 & 0.74 & 0.73 & 0.75 \\
Nemotron-Nano-9B& 0.45 & 0.61 & 0.52 & 0.67 \\
\bottomrule
\end{tabular}
\caption{Layer 2 (Analytical Agent) performance on synthetic queries ($N=2,470$). Precision, Recall, and F1 measure function-call correctness; Value Accuracy measures numerical output correctness ($\pm 1\%$ tolerance). Results shown are micro-averaged.}
\label{tab:synthetic_results}
\end{table}

\subsection{Results on Real-World Queries}

Real-world queries ($N=1,710$) introduce natural language ambiguity and require the full three-layer pipeline. We evaluate each layer separately to localize performance bottlenecks.

\paragraph{Layer 1: Input Processor.}
Table~\ref{tab:layer1_results} reports classification performance on the feasibility decision. All models achieve high recall ($>0.95$), indicating they rarely reject valid queries. Gemini 3.0 Pro achieves the best overall F1 (0.97), making it an effective gatekeeper that minimizes both false rejections and false acceptances.

\begin{table}[!htb]
\centering
\small
\begin{tabular}{l c c c c}
\toprule
\textbf{Model} & \textbf{Acc} & \textbf{Prec} & \textbf{Rec} & \textbf{F1} \\
\midrule
GPT-5.2 & 0.92 & 0.90 & 1.00 & 0.94 \\
GPT-5-Mini & 0.91 & 0.90 & 0.97 & 0.94 \\
Gemini 3.0 Pro & \textbf{0.96} & \textbf{0.95} & \textbf{1.00} & \textbf{0.97} \\
Gemini 3.0 Flash & 0.95 & 0.93 & 1.00 & 0.96 \\
Llama-4-17B & 0.86 & 0.87 & 0.95 & 0.91 \\
\bottomrule
\end{tabular}
\caption{Layer 1 (Input Processor) performance on feasibility classification ($N=1,710$). Positive class is ``answerable.'' Best results in \textbf{bold}.}
\label{tab:layer1_results}
\end{table}

\paragraph{Layer 2: Analytical Agent.}
To isolate analytical reasoning from upstream errors, we evaluate Layer 2 using only the queries correctly classified as answerable by the best Layer 1 model (Gemini 3.0 Pro), yielding $N=1,197$ samples.

Table~\ref{tab:layer2_user_results} presents the results. Performance drops substantially compared to synthetic queries: F1 falls from $\approx 0.80$ to $\approx 0.65$, and Precision drops from $\approx 0.90$ to $\approx 0.65$. This gap reflects the challenge of mapping natural language to precise function calls when user intent is ambiguous.

However, Value Accuracy remains relatively high ($>0.82$ for top models), indicating that when the agent correctly identifies the intended analysis, execution is reliable. This suggests the primary bottleneck is intent disambiguation, not computation.

\begin{table}[!htb]
\centering
\small
\begin{tabular}{l c c c c}
\toprule
\textbf{Model} & \textbf{Prec} & \textbf{Rec} & \textbf{F1} & \textbf{Val Acc} \\
\midrule
GPT-5.2 & \textbf{0.65} & \textbf{0.76} & \textbf{0.70} & 0.82 \\
GPT-5-Mini & 0.56 & 0.68 & 0.62 & 0.86 \\
Gemini 3.0 Pro & \textbf{0.65} & 0.62 & 0.64 & \textbf{0.88} \\
Gemini 3.0 Flash & \textbf{0.65} & 0.58 & 0.61 & 0.86 \\
Llama-4-17B & 0.44 & 0.66 & 0.53 & 0.44 \\
\bottomrule
\end{tabular}
\caption{Layer 2 (Analytical Agent) performance on real-world queries ($N=1,197$). Evaluation uses answerable queries from the best Layer 1 model. Results shown are micro-averaged.}
\label{tab:layer2_user_results}
\end{table}

\paragraph{Framework Extensibility.}
To validate that the Analytical Agent generalizes beyond CGM data, we conducted a pilot study integrating insulin and carbohydrate logs as additional modalities (Appendix~\ref{app:multimodal}). Evaluated on 132 new queries with the AZT1D dataset spanning single-feature extraction and multi-feature interactions using Gemini 3.0 Flash, the agent achieves F1 of 0.987 and Value Accuracy of 0.998, demonstrating that the tool-call interface supports new data modalities with minimal overhead.

\subsection{Response Quality}

Beyond numerical accuracy, we assess the readability of responses generated by Gemini 3.0 Flash given the relatively lower complexity of this task (Layer 3).

\paragraph{Qualitative Examples.}

Representative outputs for both answerable and unanswerable queries in \S\ref{sec:response_generator} demonstrate the agent's ability to (1) transform indirect questions into data-driven insights, and (2) provide informative refusals that explain limitations and suggest alternatives.

\paragraph{Readability Analysis.}
We assessed linguistic quality of all generated responses ($N=1,710$) using standard readability metrics \citep{kincaid1975derivation}. Table~\ref{tab:readability} summarizes the results.

\begin{table}[t]
\centering
\small
\begin{tabular}{@{}l c l@{}}
\toprule
\textbf{Metric} & \textbf{Value} & \textbf{Interpretation} \\
\midrule
Avg.\ Length & 108 words & Concise paragraph \\
Flesch Reading Ease & 60.3 & Standard English \\
Flesch-Kincaid Grade & 9.7 & 10th grade level \\
\bottomrule
\end{tabular}
\caption{Readability analysis of generated responses ($N=1,710$). Scores indicate accessible, non-technical language appropriate for most CGM users.}
\label{tab:readability}
\end{table}

The Flesch-Kincaid Grade of 9.7 corresponds to a 10th-grade reading level. Although this exceeds the general AMA recommendation of 6th-8th grade \cite{eltorai2014readability}, the score is artificially inflated by essential multi-syllabic terms (e.g., \textit{hypoglycemia}), while the concise length (avg. 108 words) ensures readability on mobile interfaces.

\subsection{Error Analysis}

Quantitative metrics on real-world queries are lower than on synthetic ones, but manual inspection reveals that standard metrics may \textit{underestimate} the agent's true utility due to inherent ambiguity. We identify two primary sources of valid divergence.

\paragraph{Intent Ambiguity.}
Real-world queries often underspecify the desired analysis. A question like \textit{``How was my blood glucose yesterday?''} can be validly answered in multiple ways:
\begin{itemize}[itemsep=2pt, parsep=0pt, topsep=0pt]
    \item \textbf{Aggregate metrics:} Report TIR, mean glucose, or variability for the day.
    \item \textbf{Event detection:} Identify specific excursions (e.g., ``You had a sharp drop of 45 mg/dL at 3 PM'').
    \item \textbf{Visualization:} Generate a trend plot showing the full day's trajectory.
\end{itemize}
Our ground truth captures only one interpretation per question, established through consensus discussions among the authors, including diabetes domain experts, to standardize ambiguous intents. If the agent produces a valid alternative (e.g., event detection instead of aggregate metrics), it is penalized despite providing a potentially useful answer for some users depending on their (unobserved) true informational needs. This inflates Precision/Recall errors.

\paragraph{Temporal Ambiguity.}
Implicit time references like \textit{``this afternoon''} lack precise boundaries. If the ground truth defines ``afternoon'' as 12:00--17:00 but the agent interprets it as 13:00--18:00, numerical outputs will differ even though both interpretations are reasonable. This inflates Value Accuracy errors despite correct reasoning.

\paragraph{Robustness to Glucose Variability.}
We also examine whether performance differences across subjects are driven by underlying glucose variability rather than query difficulty. To investigate this, we compute Pearson correlations between each subject's TIR and their F1 and Value Accuracy scores under Gemini 3.0 Pro. Neither correlation is statistically significant (see Appendix~\ref{app:tir_correlation}), suggesting that measured errors may primarily reflect linguistic ambiguity rather than systematic sensitivity to clinical data characteristics.

\paragraph{Implications.}
These findings motivate Layer 3's response explicitly stating the time period analyzed (e.g., \textit{``Based on your data from 12 PM to 5 PM...''}). This transparency allows users to verify whether the agent's interpretation matches their intent, mitigating the practical impact of such ambiguities.

\subsection{Ablation Study}
To quantify the contribution of the Layer 1 Input Processor, we conducted an ablation study by removing this module. 

\paragraph{Setup.} 
We constructed an experiment where raw user queries were fed directly into the Analytical Agent (Layer 2) alongside the timestamp context based on the CGM data used (e.g., \textit{``Today is 2024-03-15. User Question: ...''}). We evaluated this baseline on the same subset of answerable questions ($N \approx 1197$) used in the main experiment to ensure a fair comparison.

\paragraph{Results.} 
Table \ref{tab:ablation} isolates the contribution of the \textit{Input Processor} (Layer 1) using Gemini 3.0 Pro. The results demonstrate that the full pipeline substantially outperforms the direct-to-agent baseline across all metrics.
\begin{itemize}[itemsep=2pt, parsep=0pt, topsep=0pt]
\item \textbf{Impact on Feature Instantiation (F1 Score):} The significant decline in F1 ($\Delta=0.19$) reveals that the primary bottleneck for raw LLMs is correctly \textbf{identifying and aligning} the required feature parameters. Without the Input Processor, the model frequently omits necessary arguments or hallucinates invalid feature keys.

\item \textbf{Latent Tool-Use Capability (Value Acc.):} Value Accuracy remains relatively high (0.82). This suggests that the failure mode is parsing question intent, not reasoning. Once the correct features are identified, the agent is highly effective at deriving the correct answer. This validates the practical utility of the tool-use paradigm, provided that the input arguments are rigorously grounded.
\end{itemize}
These findings underscore our design choice: decoupling semantic refinement from computational execution is essential for handling real-world ambiguity. As shown 
in Appendix~\ref{app:interactive}, this design naturally extends to \textbf{interactive clarification} on Input Processor, yielding further performance gains on ambiguous queries.

\begin{table}[!htb]
\centering
\small
\renewcommand{\arraystretch}{1.2}
\begin{tabular}{lcc} 
\toprule
\textbf{Configuration} & \textbf{F1 Score} & \textbf{Value Acc.} \\
\midrule
\textbf{Ours (Full Pipeline)} & \textbf{0.64} & \textbf{0.88} \\
w/o Input Processor & 0.45 & 0.82 \\
\midrule
\textbf{Performance Drop ($\Delta$)} & \textit{-0.19} & \textit{-0.06} \\
\bottomrule
\end{tabular}
\caption{Ablation results on the answerable user query subset ($N=1,197$) using Gemini 3.0 Pro. Removing the Input Processor leads to significant degradation, particularly in F1 Score, highlighting the necessity of explicit query instantiation.}
\label{tab:ablation}
\end{table}

%% file: 6_conclusion.tex
\section{Conclusion} \label{sec:conclusion}

Our results in this work suggest that privacy and utility in patient-facing health agents need not be inherently in conflict. CGM-Agent serves as a real-world case study in diabetes care, showing that a carefully designed interface between natural language queries and clinically grounded computation can support meaningful patient interactions without exposing raw data. Across both our scalably parameterized questions and ambiguous user-grounded questions, existing LLMs perform reliably when operating this interface. We release CGM-Agent and our benchmark as a foundation for future work, and invite the community to join in building systems to better support patient-facing health AI.

%% file: 7_limitations_and_ethics.tex
\section*{Limitations}

While our modular framework supports integration of multiple data modalities, real-world diabetes management involves a broader range of signals not yet validated in our system, such as physical activity and sleep. Extending the framework to incorporate these modalities remains an important direction for future work, and will depend on the availability of large-scale, naturalistic multimodal datasets.

The benchmark ground truth reflects our best assessment of the most valid reference interpretation per question, yet some natural language queries admit multiple defensible analyses given the same data. Our error analysis suggests this ambiguity accounts for a meaningful portion of measured errors, and developing useful frameworks for rewarding semantically defensible alternatives could be a fruitful pathway to metrics that account for this.

Given the specialized population (CGM users) and challenge of obtaining realistic questions in-situ, we opted to conduct in-depth interviews with a modest number of participants rather than quick surveys with a broader set. We chose this approach to surface qualitatively rich, ecologically grounded queries, which we believe is a strength of this work. The interview protocol is lightweight and can be repeated to expand the dataset in future work.

\section*{Ethical Considerations}
\paragraph{Clinical Safety and Scope.} Our system focuses on retrospective analysis and reflection rather than prospective recommendations. Given the clinical sensitivity of diabetes management and associated ethical considerations, we deliberately avoid generating behavioral suggestions (e.g., insulin dosing adjustments) that could pose safety risks without professional oversight. Extending the framework to support safe, clinically validated recommendations remains an important direction for future work.

\paragraph{Data Privacy and Participant Protection.} Our user study protocol prioritized data minimization and was approved by Committee for Protection of Human Subjects at Dartmouth College. We did not collect raw sensor data from participants; data collection was restricted exclusively to written, free-form textual questions. To mitigate the risk of disclosing Personally Identifiable Information (PII), all collected queries were manually reviewed and de-identified prior to inclusion in the dataset. All data was stored on secure, HIPAA-compliant servers in accordance with our approved protocol.

%% file: 8_appendix.tex
\section{Appendix: CGM Analysis Toolkit}
\label{sec:appendix_toolkit}

Table~\ref{tab:full_tool_definitions} lists the complete set of deterministic functions implemented in our toolkit. These functions are categorized into three groups: 
\begin{itemize}
    \item \textbf{Data Processing \& CGM weartime}: Handles data ingestion, filtering, and quality checks.
    \item \textbf{Daily Metrics Extraction}: Calculates clinical metrics for individual days (e.g., TIR, GMI).
    \item \textbf{Long-term Aggregation \& Analysis}: Performs statistical operations across multiple days (e.g., averages, trends, comparisons). 
\end{itemize}

\begin{table*}[t!]
\centering
\small
\renewcommand{\arraystretch}{1.3}

\caption{Complete definition of Python tools available to the Agentic Framework.}
\label{tab:full_tool_definitions}

\begin{tabular}{p{2.0cm} p{5.1cm} p{7.9cm}}
\toprule
\textbf{Category} & \textbf{Function Name} & \textbf{Description \& Clinical Utility} \\
\midrule

\multirow{3}{*}{\parbox{2.5cm}{Data Processing\\\& CGM Weartime}} 
 & \texttt{filter\_cgm\_csv} & 
 Filters raw CGM data based on a user-specified date list or time window (e.g., "last 3 days" or "6AM-12PM"). Handles chronological sorting. \\[6pt]
 
 & \texttt{estimate\_cgm\_sampling\_rate} & 
 Infers the device sampling rate (e.g., 5 min vs. 15 min) to accurately calculate expected readings for CGM weartime checks. \\[6pt]
 
 & \texttt{find\_adherence} & 
 Calculates the percentage of active wear time for full days or specific time windows to determine data sufficiency. \\
\midrule

\multirow{5}{*}{\parbox{2.5cm}{Daily Clinical\\Metrics}} 
 & \texttt{find\_BG\_time\_range} & 
 Computes percentage and duration of Time in Range (TIR), Time below Range (TBR), and Time above Range (TAR). \\[6pt]
 
 & \texttt{find\_avg\_std\_gv\_BG} & 
 Calculates Mean blood glucose, Standard Deviation, Glycemic Variability (CV), estimates A1c and  Glucose Management Indicator (GMI) based on mean glucose. \\[6pt]
 
 & \texttt{find\_BG\_min\_max} & 
 Identifies the absolute minimum and maximum glucose values for specified dates to detect extreme outliers. \\[6pt]
 
 & \texttt{find\_hypo\_hyper\_events} & 
 Counts discrete events of Hypoglycemia ($<70$ mg/dL for 15+ min) and Hyperglycemia ($>180$ mg/dL for 15+ min). \\[6pt]
 
 & \texttt{extract\_features\_json} & 
 Pipeline wrapper that executes all daily metric functions and aggregates results into a structured JSON for downstream analysis. \\
\midrule

\multirow{6}{*}{\parbox{2cm}{Long-term\\Aggregation\\\& Analysis}} 
 & \texttt{get\_average} & 
 Computes the average of a feature (e.g., TIR) across multiple days. Crucially, it returns two values: one for "All Data" and one specifically for "Good Weartime" ($\ge70\%$) days. \\[6pt]
 
 & \texttt{count\_satisfied\_condition} & 
 Counts how many days meet a specific criterion (e.g., "Days with no hypoglycemia events"). \\[6pt]
 
 & \texttt{feature\_range} & 
 Finds the global minimum and maximum of a specific feature (e.g., "Which day had the lowest TIR?") across a long period. \\[6pt]
 
 & \texttt{compute\_difference\_ratio} & 
 Compares two different time periods (e.g., "Last week vs. This week") and calculates the absolute difference and ratio for any feature. \\[6pt]
 
 & \texttt{calculate\_blood\_glucose\_excursion} & 
 Detects rapid glycemic excursions (spikes/drops). \\[6pt]
 
 & \texttt{plot\_daily\_trends} & 
 Generates and saves a 24-hour aggregate plot (Average Daily Profile) to visualize daily trends and patterns. see Figure \ref{fig:trend_example})\\

\bottomrule
\end{tabular}
\end{table*}

\begin{figure}[h]
    \centering
    \includegraphics[width=\linewidth]{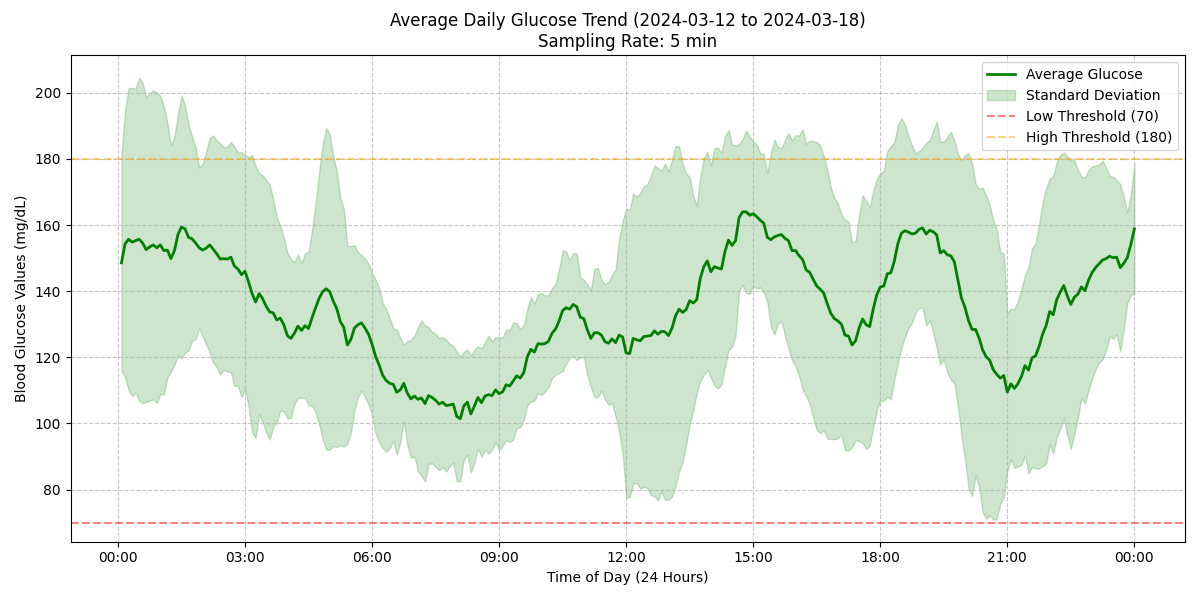}
    \caption{\textbf{Average Daily Glucose Profile.} Generated by the \texttt{plot\_daily\_trends} function, this visualization aggregates 7 days of data to display the mean glucose trajectory (solid green line) and glucose standard deviation (shaded area, $\pm$1 SD) relative to standard clinical target boundaries (70--180 mg/dL).}
    \label{fig:trend_example} 
\end{figure}

\section{Appendix: Control flow for two representative query types.}
\label{sec:control_flow_appendix}

\begin{figure*}[t]
    \centering
    \includegraphics[width=0.94\textwidth]{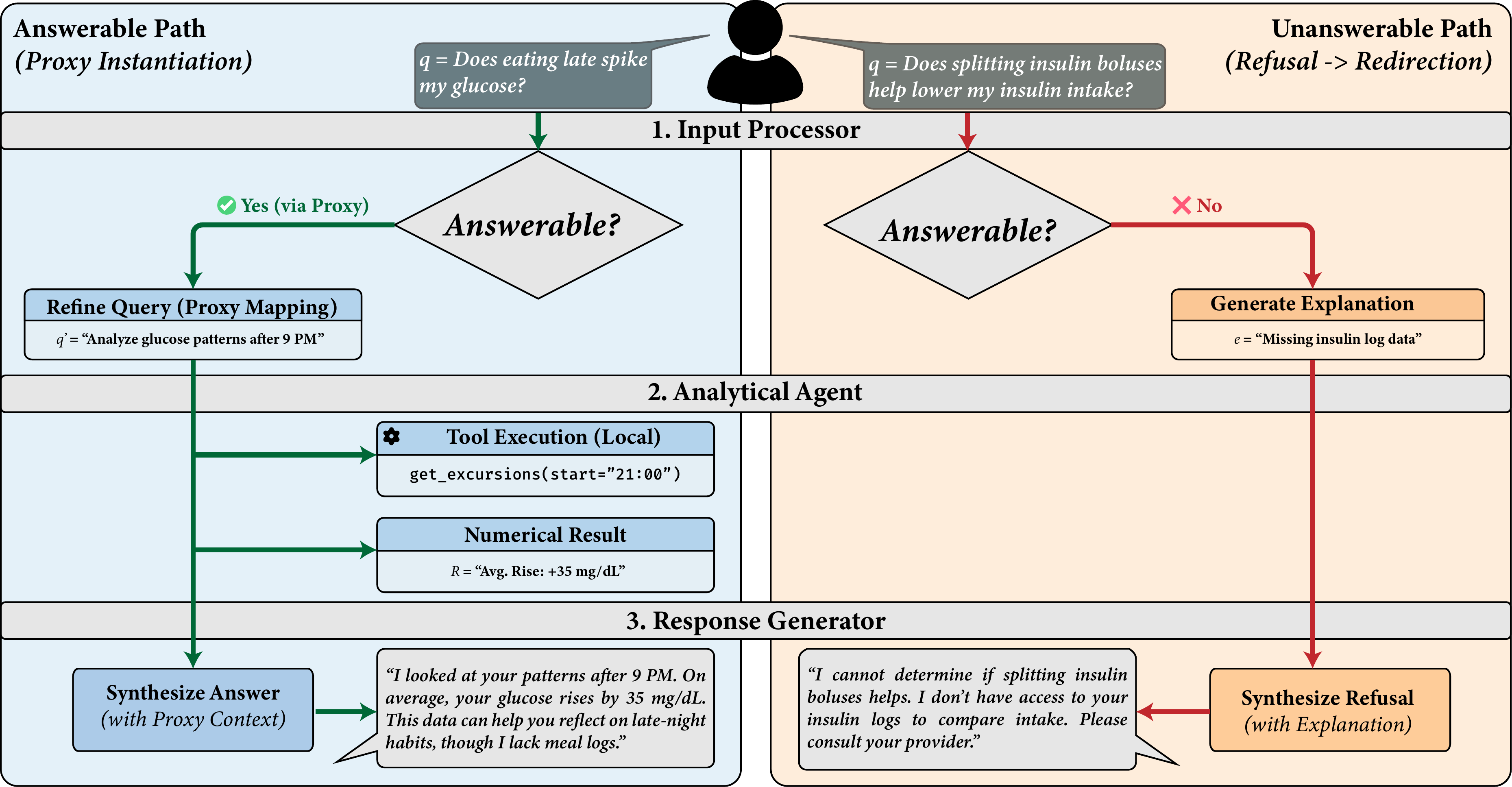}
    \caption{Control flow for two representative query types. \textbf{Left (blue):} A query referencing unobserved behavior (late-night eating) is mapped to a proxy instantiation and answered with a reflection-prompting response. \textbf{Right (orange):} A query requiring external data (insulin logs) is classified as unanswerable; Layer 2 is bypassed and the system generates an explanation.}
    \label{fig:control_flow}
\end{figure*}

Figure~\ref{fig:control_flow} shows that a query that can be approximately answered via proxy instantiation, and an unanswerable query requiring graceful refusal.

\section{Interactive Clarification Study}
\label{app:interactive}

\subsection{Setup}
The Input Processor (Layer~1) can ask a clarification question upon detecting ambiguity, rather than defaulting to a fixed interpretation. This requires no architectural changes---it is a natural extension of the existing query refinement logic in Layer~1.

We evaluated this \textit{Interactive} clarify-then-answer setting against the \textit{Original} single-turn setting on 63 ambiguous real-user questions per subject across four subjects (2~T1D, 2~T2D) using Gemini~3~Pro. A simulated user agent, with access to the underlying data templates, provides grounded responses to clarification questions.

\subsection{Results}

\begin{table}[h]
\centering
\small
\begin{tabular}{llcccc}
\toprule
\textbf{Subject} & \textbf{Setting} & \textbf{Prec} & 
\textbf{Rec} & \textbf{F1} & \textbf{Val Acc} \\
\midrule
P1  & Original    & 0.65 & 0.59 & 0.62 & 0.91 \\
  & Interactive & \textbf{0.67} & \textbf{0.77} & 
    \textbf{0.71} & \textbf{0.93} \\
\midrule
P11 & Original    & 0.60 & \textbf{0.57} & \textbf{0.58} 
    & 0.79 \\
 & Interactive & \textbf{0.64} & 0.42 & 0.51 & 
    \textbf{0.87} \\
\midrule
P12 & Original    & 0.67 & 0.54 & 0.60 & 0.82 \\
 & Interactive & \textbf{0.73} & \textbf{0.77} & 
    \textbf{0.75} & \textbf{0.86} \\
\midrule
P19 & Original    & 0.63 & 0.59 & 0.61 & 0.88 \\
 & Interactive & \textbf{0.63} & \textbf{0.69} & 
    \textbf{0.66} & \textbf{0.91} \\
\midrule
Avg & Original    & 0.64 & 0.57 & 0.60 & 0.85 \\
 & Interactive & \textbf{0.67} & \textbf{0.66} & 
    \textbf{0.66} & \textbf{0.89} \\
\bottomrule
\end{tabular}
\caption{Interactive clarification results on 63 ambiguous real-user questions per subject (Gemini~3~Pro). \textbf{Bold} marks the higher-performing setting within each subject pair.}
\label{tab:interactive}
\end{table}

A single clarification round improves performance for 3 out of 4 subjects in F1, and all 4 subjects in Value Accuracy, yielding average absolute gains of $+5.5\%$ F1 and $+4.3\%$ Value Accuracy. For P12, the F1 score increased significantly from 0.60 to 0.75, which suggests that the agent effectively synthesizes user-provided constraints to refine its analysis. For P11, clarification improved accuracy but led to more conservative tool selection (higher precision, lower recall), suggesting the clarification sometimes narrowed scope excessively.

\subsection{Clarification Example}
The following exchange illustrates how the Input Processor resolves temporal ambiguity through clarification:

\begin{quote}
\textbf{User:} \textit{``What is the standard deviation of my blood glucose around dawn?''} \\[4pt]
\textbf{Agent:} \textit{``Please specify the time range you consider `dawn' and the dates you would like analyzed.''} \\[4pt]
\textbf{User:} \textit{``4~AM to 6~AM on 2024-02-29.''} \\[4pt]
\textbf{Refined query with interaction:} \textit{``What are my standard deviation of blood glucose and CGM adherence for the time range \textbf{2024-02-29 04:00 to 2024-02-29 06:00?}''} \\[4pt]
\textbf{Refined query without interaction:} \textit{``What is the standard deviation of my blood glucose restricted to the time range \textbf{04:00 to 07:00 (dawn)} over the following dates: 2024-01-01 to 2024-02-29? Also provide the CGM adherence for these dates.``}
\end{quote}

\section{Appendix: Multi-modal Extensibility Pilot}
\label{app:multimodal}

To validate the modular design of the Analytical Agent, we integrated insulin and carbohydrate logs as additional data sources alongside CGM data, using the T1D cohort (AZT1D dataset). New local executors were registered for each modality without modifying the core framework.

We evaluated the agent on 132 new queries in two categories:
\begin{itemize}[itemsep=2pt, parsep=0pt, topsep=0pt]
    \item \textbf{Single-Feature Extraction} (66 queries): Calculating specific daily totals for insulin or carbohydrate intake.
    \item \textbf{Multi-Feature Interaction} (66 queries): Analyzing glucose fluctuations within 1--3 hours following recent insulin doses or carbohydrate intake.
\end{itemize}

\begin{table}[h]
\centering
\small
\begin{tabular}{l c c c c}
\toprule
\textbf{Model} & \textbf{Prec} & \textbf{Rec} & \textbf{F1} & \textbf{Val Acc} \\
\midrule
Gemini 3.0 Flash & 0.996 & 0.978 & 0.987 & 0.998 \\
\bottomrule
\end{tabular}
\caption{Agent performance on multi-modal queries integrating insulin and carbohydrate logs ($N=132$). Results confirm that adding new data modalities requires only registering a corresponding local executor.}
\label{tab:multimodal}
\end{table}

The system achieves near-ceiling performance on both query types, confirming that the tool-call interface generalizes cleanly to new modalities. Full integration of physical activity, sleep, and meal logs remains an important direction for future work, and will depend on the availability of large-scale naturalistic multimodal datasets.

\section{Appendix: Robustness to Glucose Variability}
\label{app:tir_correlation}

To assess whether agent performance is confounded by underlying glucose variability, we compute Pearson correlations between each subject's TIR and two performance metrics (F1 and Value Accuracy) using Gemini 3.0 Pro results. Results are reported overall and by diabetes type in Table~\ref{tab:tir_correlation}.

\begin{table}[h]
\centering
\small
\begin{tabular}{l l r r}
\toprule
\textbf{Group} & \textbf{Metric Pair} & \textbf{$r$} & \textbf{$p$-value} \\
\midrule
Overall & TIR vs.\ F1       & 0.385 & 0.104 \\
 & TIR vs.\ Val Acc  & 0.095 & 0.699 \\
\midrule
T1D     & TIR vs.\ F1       & 0.358 & 0.280 \\
   & TIR vs.\ Val Acc  & $-$0.044 & 0.898 \\
\midrule
T2D     & TIR vs.\ F1       & 0.622 & 0.100 \\
    & TIR vs.\ Val Acc  & 0.181 & 0.667 \\
\bottomrule
\end{tabular}
\caption{Pearson correlation between subject-level TIR and agent performance metrics (Gemini 3.0 Pro).}
\label{tab:tir_correlation}
\end{table}

No correlation reaches statistical significance across any group. The moderate coefficient in the T2D subgroup ($r = 0.622$) does not survive the significance threshold ($p = 0.100 > 0.05$), likely reflecting the small subgroup size ($N = 8$) rather than a genuine performance dependency. These results suggest that the agent's accuracy is not systematically driven by the glucose management in the underlying data.

\section{Appendix: User Question Analysis and Instantiation}
\label{sec:appendix_questions}
As discussed in \S\ref{sec:dataset}, real-world user queries may be ambiguous and beyond simple statistical retrieval. To handle this, our Input Processor (Layer 1) classifies questions based on feasibility and maps subjective intent to computable proxies.

\subsection{Unanswerable Questions}
Table \ref{tab:unanswerable_examples} lists examples of queries classified as Unanswerable. These typically fall into categories such as:
\begin{itemize}
    \item \textbf{Missing Modality:} Requests requiring external data logs not present in standard CGM files (e.g., insulin type, food ingredients, activity and sleep logs).
    \item \textbf{Predictive/Causal Inference:} Requests asking for future predictions or definitive causal links (e.g., "Did X cause Y?") which cannot be clinically proven solely from retrospective glucose data.
    \item \textbf{Medication Consultation:} Queries seeking specific medical advice, drug pharmacodynamics, or long-term prognosis, which require professional clinical judgment beyond the scope of a data analysis agent.
\end{itemize}

\begin{table*}[h]
\centering
\small
\renewcommand{\arraystretch}{1.3}
\begin{tabular}{p{0.45\textwidth} p{0.45\textwidth}}
\toprule
\textbf{User Query ($q$)} & \textbf{Reason for Rejection / Missing Modality} \\
\midrule
\textit{``What type of foods are generally safe to consume and do not need massive insulin boluses?''} & \textbf{Missing Dietary Log:} CGM data contains glucose levels but lacks specific food intake records required to correlate food types with insulin response. \\[6pt]

\textit{``Does splitting insulin boluses help me lower my insulin intake?''} & \textbf{Missing Medication Log:} Requires detailed insulin dosing logs (timing, amount, split vs. single) which are external to the CGM sensor data. \\[6pt]

\textit{``Given my CGM data, can you find moments that were likely incorrect blood glucose values?''} & \textbf{Sensor Hardware Limitation:} Detecting sensor errors (compression lows, calibration drifts) requires raw electrical signal data or external calibration values, not just the processed glucose stream. \\[6pt]

\textit{``Does the stability of my blood glucose over the last three days predict more stability in the coming days?''} & \textbf{Forecasting/Prediction:} The current agent is designed for retrospective analysis and insight generation, not predictive modeling of future physiological states. \\[6pt]

\textit{``How does Mounjaro impact insulin intake, and what are the long-term effects on insulin-to-carb ratios? Will these ratios return to what they were...?''} & \textbf{Medication Consultation:} Questions regarding specific drug mechanisms (e.g., GLP-1 agonists) and long-term medical prognosis constitute clinical advice, which is strictly out of scope for a retrospective data analysis agent. \\
\bottomrule
\end{tabular}
\caption{Examples of user questions labeled as \textbf{Unanswerable} in our dataset ($N=513$). The Input Processor is trained to generate an explanation $e$ (right column) instead of attempting a hallucinated calculation.}
\label{tab:unanswerable_examples}
\end{table*}

\subsection{Proxy Instantiations}
For questions that imply a causal link (e.g., impact of lifestyle) but lack explicit logs, we employ Proxy Instantiation. As shown in Table \ref{tab:proxy_mapping}, we map these intent-driven questions into time-based comparative analyses. This allows the agent to provide \textit{data-driven correlations} (e.g., "Glucose was lower during your workout window") while acknowledging the lack of definitive causation.

\begin{table*}[h]
\centering
\small
\renewcommand{\arraystretch}{1.3}
\begin{tabular}{p{0.45\textwidth} p{0.45\textwidth}}
\toprule
\textbf{Ambiguous User Query ($q$)} & \textbf{Instantiated Computable Proxy ($q'$)} \\
\midrule
\textit{``Given my CGM data, can you infer when I ate my meals?''} & \textbf{Trend Analysis:} ``Given my CGM data over [Date], can you infer when my glucose level rises fast?''  \\[6pt]

\textit{``What is my time in range during workouts vs. after exercise?''} & \textbf{Temporal Windowing:} ``What is my time in range between [Datetime A] and [Datetime B] ?''  \\[6pt]

\textit{``How long after ingesting food do my glucose levels rise?''} & \textbf{Event Detection:} ``When my glucose levels rise over today?''  \\[6pt]

\textit{``What patterns do I see around my menstrual cycle?''} & \textbf{Blood Glucose Trends:} ``Plot my typical daily CGM blood glucose trends for the following dates: [Date A] to [Date B].'' \\
\bottomrule
\end{tabular}
\caption{Examples of \textbf{Proxy Instantiations} ($N=399$). The vague user intent (left) is transformed into a precise, computable query ($q'$) targeting specific time windows or statistical features (right), enabling the Analytical Agent to execute code.}
\label{tab:proxy_mapping}
\end{table*}

\section{Appendix: System Prompts}
\label{sec:appendix_prompts}

We employ distinct system instructions for each layer of the CGM-Agent framework. We present the core prompts used for the Input Processor, Analytical Agent, Response Generator, and the Automated Judgment.

\subsection{Agent Prompts}
\label{sec:appendix_agent_prompts}

\begin{promptbox}{Layer 1: Input Processor (Query Refiner)}
\textbf{Role:} You are a Question Refiner that processes raw user questions about CGM data into a standardized, answerable format for a data analysis agent.\\

\textbf{Goal:} Determine if the question can be answered using only CGM glucose timestamp and value data, and if so, rephrase it into a standard query.\\

\textbf{Supported Features:}
1. Time in Range (TIR), Time below Range (TBR), Time above Range (TAR), ideal blood glucose control \\
2. Average blood glucose, Standard Deviation, Glycemic Variability (CV), estimated A1c, estimated Glucose Management Indicator (eGMI) \\
3. Min/Max blood glucose, Hypoglycemia and hyperglycemia events \\
4. Glucose excursions and glucose trends \\

\textbf{Answerability Logic:}
1. \textbf{Direct Data (YES):} Questions about past glucose data.\\
2. \textbf{Behavioral (Indirect YES):} Questions about "food/exercise/sleep" ARE answerable IF convertible to glucose trends during a specific time.\\
3. \textbf{Medical/External (NO):} General medical knowledge, future predictions, or questions strictly requiring insulin/food logs (e.g., "What is my insulin sensitivity?").\\

\textbf{Refinement Guidelines (Standardized Formats):}
While extracting features, consider if user needs to know CGM weartime to determine if the calculation includes enough data points to show that feature is reliable. Some standardized formats to make your refined questions be clear with specific information:
\begin{itemize}
    \item \textbf{Basic Retrieval:} ``What are my \{features\} and CGM weartime over the following dates: \{dates\_str\}?''
    \item \textbf{Conditional Statistics:} ``What are my average \{features\} over \{dates\_str\}? Consider two conditions: 1. Days with any CGM records. 2. Days with good weartime (>70\%).''
    \item \textbf{Cohort Comparison:} ``Compare \{features\} between groups of dates: \{dates\_str\}. Calculate: 1. Average value per group. 2. Absolute difference. 3. Which group is higher.''
    \item \textbf{Event Analysis:} ``Analyze glucose excursions for \{dates\_str\}. Find significant rapid changes and details on timing, magnitude, and speed.''
    \item \textbf{Visualization:} ``Plot my typical daily CGM blood glucose trends for \{dates\_str\}. Output the mean values used to generate the plot.''
\end{itemize}

\textbf{Input Fields:}
\begin{itemize}
    \item \texttt{user\_question}: The original raw question.
    \item \texttt{reference\_date}: To interpret "yesterday", "last week".
    \item \texttt{reference\_datetime}: To interpret "over last 4 hours".
\end{itemize}

\textbf{Output Fields:}
\begin{itemize}
    \item \texttt{is\_answerable}: Boolean.
    \item \texttt{refined\_question}: The standardized query or specific missing modality explanation.
    \item \texttt{rationale}: Reasoning for how agent process the question.
\end{itemize}
\end{promptbox}


\begin{promptbox}{Layer 2a: Analytical Router (Task Decomposer)}
\textbf{Role:} You are a Router Agent that acts as the entry point for the analytical pipeline.\\

\textbf{Goal:} Analyze the user's request to determine if it constitutes a \textbf{Single Task} or \textbf{Multiple Separate Tasks}, and delegate accordingly.\\

\textbf{Routing Logic:}\\
1. \textbf{Single Task (Batch/Comparison):}\\
   - \textit{Pattern:} Requests involving a specific list of dates (e.g., "Dec 29, Dec 31") or explicit comparisons (keywords: "compare", "difference", "vs").\\
   - \textit{Action:} Treat as ONE self-contained task. Do NOT split.\\
   - \textit{Example:} "Compare TIR between Group A dates and Group B dates." $\rightarrow$ Send as 1 query.

2. \textbf{Multiple Tasks (Decomposition):}\\
   - \textit{Pattern:} Requests containing keywords like "separately", "each week", or distinct disjoint time ranges (e.g., "Week 1 AND Week 2").\\
   - \textit{Action:} Split into a list of focused sub-questions to be executed independently.\\
   - \textit{Example:} "What was my average glucose for Week 1 and Week 2 separately?" $\rightarrow$ Split into ["Average glucose Week 1?", "Average glucose Week 2?"].\\

\textbf{Input:} \texttt{user\_request}: Refined question from Input Processor. 

\textbf{Output:}
\begin{itemize}
    \item \texttt{date\_list}: List of date strings/ranges.
    \item \texttt{question\_list}: List of decomposed queries to be sent to the Executor.
\end{itemize}
\end{promptbox}

\begin{promptbox}{Layer 2b: Analytical Executor}
\textbf{Role:} You are a Healthcare Scientist and the primary worker agent.

\textbf{Workflow:} You MUST strictly follow this three-step pipeline:
\texttt{filter\_cgm\_csv} $\rightarrow$ \texttt{extract\_features\_json} $\rightarrow$ \texttt{computation\_tool}.\\

\textbf{Execution Constraints:}
\begin{itemize}
    \item \textbf{Internal Data Only:} Process features derivable strictly from CGM timestamps and values.
    \item \textbf{Missing Modalities:} If a feature requires external logs (insulin, food, sleep), set value to \textbf{-1}.
    \item \textbf{Boolean Logic:} For Yes/No questions, return \textbf{1} (Yes) or \textbf{0} (No).
    \item \textbf{No Hallucination:} Do not invent features not present in the data.
\end{itemize}

\textbf{Output Format Schema:}
Return a nested dictionary: \texttt{\{Date\_Key: \{Feature\_Name: Value\}\}}.
\begin{itemize}
    \item \textbf{Single Date:} \texttt{"2025-09-01": \{"TIR": 70.5\}}
    \item \textbf{Date Range:} \texttt{"(2025-09-01, 2025-09-07)": \{"mean\_glucose": 120\}}
    \item \textbf{Date List:} \texttt{"['2025-01-01', '2025-01-03']": \{"days\_with\_good\_weartime": 2\}}
\end{itemize}

\textbf{Input:} A single specific query from the Router. \\

\textbf{Output:} \texttt{result} (The computed numerical dictionary).
\end{promptbox}

\begin{promptbox}{Layer 3: Response Generator}
\textbf{Goal:} Generates a clear, concise, and empathetic final response to a user about their CGM (Continuous Glucose Monitor) data.

\textbf{Response Guidelines:}
\begin{enumerate}
    \item \textbf{Handling Refusals:} Since \texttt{include\_rationale} is True, if \texttt{is\_answerable} is False, you MUST explain \textit{why} based on the \texttt{rationale} (e.g., "I cannot analyze this because I lack food logs").
    \item \textbf{Answer Structure:}
        \begin{itemize}
            \item \textbf{Direct Answer:} Start with the key finding from \texttt{execution\_result}.
            \item \textbf{Contextual Bridge:} Explain how the data relates to the user's intent (e.g., "To answer your question about glucose during exercise, I looked at your glucose from 3 - 5 PM").
            \item \textbf{Data Evidence:} Cite specific numbers/trends from \texttt{execution\_result} to support the claim.
        \end{itemize}
\end{enumerate}

\textbf{Dynamic Input Fields (Per-Query):}
\begin{itemize}
    \item \texttt{raw\_question}: The user's original raw input.
    \item \texttt{is\_answerable}: Boolean flag indicating feasibility.
    \item \texttt{rationale}: Reasoning for why the question was accepted or rejected.
    \item \texttt{execution\_result}: The raw JSON output containing numerical data.
    \item \texttt{tone}: The tone of the response. 
    \item \texttt{complexity\_level}: The target audience complexity level (e.g., middle school knowledge).
\end{itemize}

\textbf{Output:} \texttt{final\_response} (Natural language text).
\end{promptbox}

\subsection{Evaluation Prompts of Analytical Agent}
\label{sec:appendix_eval_prompts}

\begin{promptbox}{Automated Evaluator (Gemini 3.0 Pro)}
\textbf{Task:} rigorous comparison between the Agent's numerical output and Ground Truth (GT).

\textbf{Feature Matching Guidelines:}
\begin{itemize}
    \item \textbf{Semantic Mapping:} Match features by concept, not exact naming (e.g., "avg bg" $\equiv$ "mean blood glucose").
    \item \textbf{Missing Data Handling:} 
    \begin{itemize}
         \item If GT is -1 (No Data) and Agent result is missing that feature $\rightarrow$ \textbf{Match} (Both correctly identified no data).
         \item If GT has a valid value but Agent is missing it $\rightarrow$ \textbf{Mismatch} (False Negative).
    \end{itemize}
\end{itemize}

\textbf{Value Comparison Logic:}
\begin{itemize}
    \item \textbf{Numerical Tolerance:} Values match if within $\pm$ 1\%.
    \item \textbf{Boolean Logic:} 1.0 = "Yes", 0.0 = "No".
    \item \textbf{Special Case (-1 vs 0):}
    \begin{itemize}
        \item \textit{CGM Weartime/Usage:} -1 and 0 are considered \textbf{Equivalent} (both imply "No Data").
        \item \textit{Event Counts/TIR:} -1 (Missing Data) $\neq$ 0 (Zero occurrences). These represent different clinical states.
    \end{itemize}
\end{itemize}

\textbf{Filtering Logic:} (Primarily for Real-world User Queries)
\begin{itemize}
    \item If \texttt{required\_features} is provided: ONLY evaluate features in this list. Ignore extras.
    \item If \texttt{required\_features} is empty: Evaluate ALL features present in GT.
\end{itemize}

\textbf{Input Fields:}
\begin{itemize}
    \item \texttt{question}: The context query.
    \item \texttt{required\_features}: List of priority metrics to verify.
    \item \texttt{gt\_res}: Ground truth dictionary.
    \item \texttt{agent\_result}: Agent prediction dictionary.
\end{itemize}

\textbf{Output Schema:}
\texttt{comparison} dictionary containing:
\begin{itemize}
    \item Counts: \texttt{num\_gt\_features}, \texttt{num\_agent\_features}, \texttt{num\_overlap}.
    \item Sets: \texttt{features\_in\_gt\_not\_in\_agent} (FN), \texttt{features\_in\_agent\_not\_in\_gt} (FP).
    \item Value matches: \texttt{feature\_value\_comparison}. 
\end{itemize}
\end{promptbox}

\section{Appendix: Latency and Resource Footprint}
\label{app:latency}

Table~\ref{tab:latency} reports end-to-end latency for the Analytical Agent (Layer~2) across 130~queries for a randomly selected subject (P2). 

\begin{table}[h]
\centering
\small
\begin{tabular}{lcccc}
\toprule
\textbf{Model} & \textbf{Mean} & \textbf{Median} & \textbf{P95} & \textbf{LLM Calls} \\
\midrule
Gemini~3~Flash & 82.8 & 74.7 & 153.7 & 8.4 \\
Gemini~3~Pro   & 100.4 & 98.4 & 148.8 & 8.0 \\
\bottomrule
\end{tabular}
\caption{Analytical Agent (Layer~2) latency over 130 queries (P2). Mean, Median, and P95 are reported in seconds. LLM Calls reports the mean number of API calls per query.}
\label{tab:latency}
\end{table}

Latency is driven primarily by sequential LLM API calls ($\sim$8 per query) and repeated file I/O for per-request metric computation. In production, pre-computing and caching data structures would substantially reduce overhead, and 
independent sub-tasks could be parallelized. For latency-sensitive deployments, lightweight SLMs could replace the backbone for lower-complexity sub-tasks without modifying the overarching framework.

\section{Appendix: Participant Recruitment and Instruction Material}
\label{sec:appendix_recruitment}

Below is the condensed text used for participant recruitment and instructions, anonymized for review purposes.

\paragraph{Ethics and Protocol}
This study was approved by the Institutional Review Board (IRB) at Dartmouth College. Participation was voluntary. All participants were required to complete a consent form indicating their willingness to participate before proceeding with the study tasks.

\paragraph{Participant Requirements and Recruitment}
We recruited adults (aged 18 and older) who actively use a continuous glucose monitor (CGM).

\paragraph{Compensation}
Participants received a \$10 USD gift card as compensation for their time upon successfully submitting 10 or more questions.

\paragraph{Task Instructions}
Participants were informed that they did not need to share their actual raw glucose data. Instead, they were asked to write free-form questions reflecting their curiosity when reviewing their CGM data. The specific instructions provided to participants is as follows:
\begin{itemize}
    \item Questions can relate to prior glucose data (e.g., from previous hours, days, weeks, or months).
    \item Questions can seek to uncover insights regarding personal glucose management and trends.
    \item The focus should be on retrospective review (looking back at data to reflect on progress) rather than real-time monitoring (e.g., ``Is my blood glucose high or low right now?'').
    \item Questions can be short or detailed; participants are encouraged to be clear and concrete.
\end{itemize}

\section{Appendix: Subject Selection} \label{app:subject-selection}

Table~\ref{tab:subject-mapping} lists the mapping between the anonymous subject identifiers used throughout this paper and the original identifiers in their respective public datasets. All subjects were selected by prioritizing the longest available CGM recording duration within each dataset, as described in \S\ref{sec:cgm-data-source}.

\begin{table}[h]
\centering
\small
\begin{tabular}{llc}
\toprule
\textbf{Subject ID} & \textbf{Original ID} & \textbf{Dataset} \\
\midrule
P1  & Subject 15 & AZT1D \\
P2  & Subject 23 & AZT1D \\
P3  & Subject 21 & AZT1D \\
P4  & Subject 20 & AZT1D \\
P5  & Subject 7  & AZT1D \\
P6  & Subject 19 & AZT1D \\
P7  & Subject 5  & AZT1D \\
P8  & Subject 13 & AZT1D \\
P9  & Subject 6  & AZT1D \\
P10 & Subject 11 & AZT1D \\
P11 & Subject 4  & AZT1D \\
\midrule
P12 & 2069 & ShanghaiT2DM \\
P13 & 2014 & ShanghaiT2DM \\
P14 & 2017 & ShanghaiT2DM \\
P15 & 2015 & ShanghaiT2DM \\
P16 & 2078 & ShanghaiT2DM \\
P17 & 2001 & ShanghaiT2DM \\
P18 & 2055 & ShanghaiT2DM \\
P19 & 2074 & ShanghaiT2DM \\
\bottomrule
\end{tabular}
\caption{Mapping of anonymous subject identifiers to original dataset identifiers. P1--P11 are drawn from AZT1D \citep{khamesian2025type} (T1D, 5-min sampling), and P12--P19 from ShanghaiT2DM \citep{zhao2023chinese} (T2D, 15-min sampling).}
\label{tab:subject-mapping}
\end{table}